\documentclass[sigconf]{acmart}
\usepackage{graphicx}
\usepackage{enumitem}
\usepackage{caption}
\usepackage{subcaption}
\newcommand\BibTeX{B\textsc{ib}\TeX}
\usepackage{amsmath}
\def\approxprop{%
  \def\p{%
    \setbox0=\vbox{\hbox{$\propto$}}%
    \ht0=0.6ex \box0 }%
  \def\s{%
    \vbox{\hbox{$\sim$}}%
  }%
  \mathrel{\raisebox{0.7ex}{%
      \mbox{$\underset{\s}{\p}$}%
    }}%
}
\patchcmd{\maketitle}{\@copyrightspace}{}{}{}
\patchcmd{\maketitle}{\@copyrightpermissionfootnoterule}{}{}{}

\AtBeginDocument{%
  \providecommand\BibTeX{{%
    \normalfont B\kern-0.5em{\scshape i\kern-0.25em b}\kern-0.8em\TeX}}}

\begin{document}

\title{Zero-Shot Learning for Joint Intent and Slot Labeling}



\author{Rashmi Gangadharaiah}
\affiliation{
\institution{AWS AI Labs}
}
\email{rgangad@amazon.com}

\author{Balakrishnan Narayanaswamy}
\affiliation{
\institution{AWS AI Labs}
}
\email{muralibn@amazon.com}


\begin{abstract}
It is expensive and difficult to obtain the large number of sentence-level intent and token-level slot label annotations required to train neural network (NN)-based Natural Language Understanding (NLU) components of task-oriented dialog systems, especially for the many real world tasks that have a large and \textit{growing} number of intents and slot types.
While zero shot learning approaches that require no labeled examples - only features and auxiliary information - have been proposed only for slot labeling, we show that one can profitably perform \textit{joint} zero-shot intent classification and slot labeling. We demonstrate the value of capturing dependencies between intents and slots, and between different slots in an utterance in the zero shot setting. We describe NN architectures that translate between word and sentence embedding spaces, and demonstrate that these modifications are required to enable zero shot learning for this task. We show a substantial improvement over strong baselines and explain the intuition behind each architectural modification through visualizations and ablation studies.
\end{abstract}



\maketitle

\section{Introduction}
\begin{figure}
\centering
\includegraphics[width=1.0\linewidth]{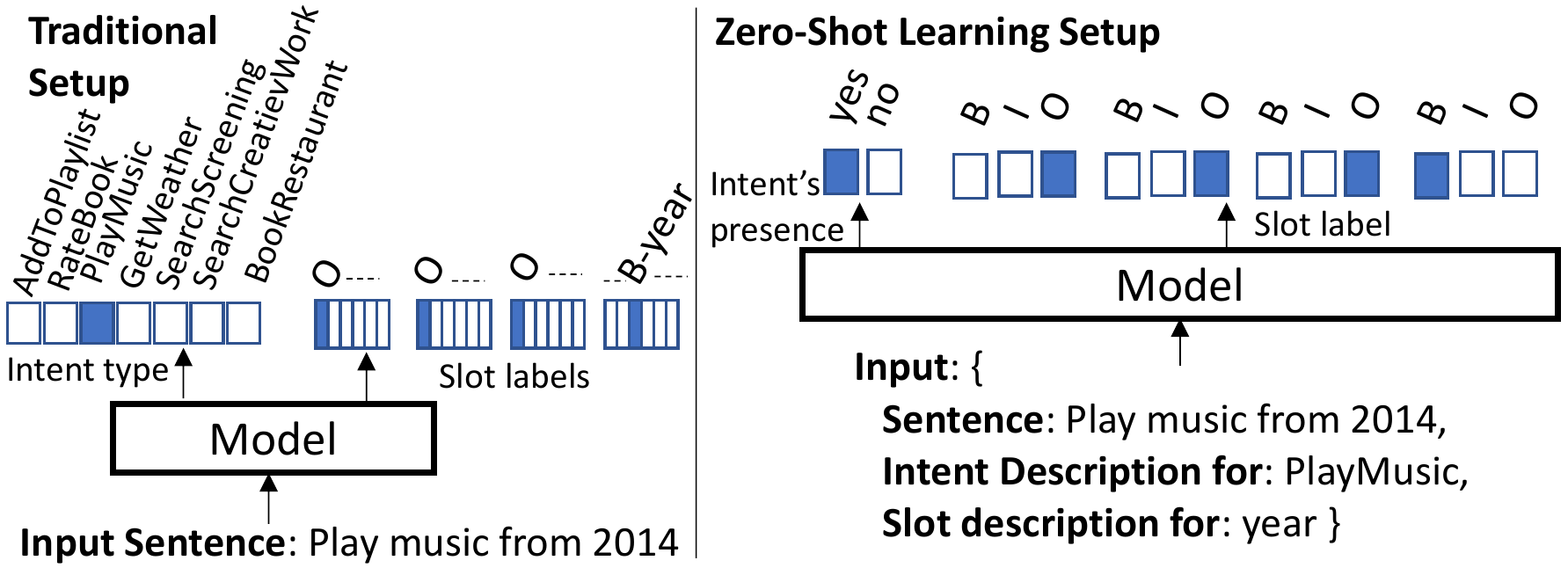}
\caption{Traditional vs. Zero-Shot learning setup. \textit{Note}: previous approaches considered either slot labeling or intent detection in zero-shot learning settings and not both.} 
\label{fig:sub2}
\end{figure}

The NLU component in task oriented dialog systems (e.g., Alexa, Google Assistant, Cortana or Siri), is responsible for identifying the user’s intent and for extracting relevant constraints from the user's utterance or input sentence. 
Consider an example in Figure \ref{fig:sub2} (see Traditional Setup). Given a user's utterance, \textit{Play music from 2014}, the NLU 
\cite{obuchowski2020transformer,goo-etal-2018-slot,BingLiu:2016,han2019delta,Zhang:2018,Qin:2019,Haihong:2019,wang-etal-2018-bi,BingLiu2:2016} detects the \textit{PlayMusic} intent and extracts slot labels relevant to this intent, such as, $\{$\textit{year=2014}$\}$. These NLU components have recently been constructed using sophisticated NN architectures \cite{Jerome:2017,Mesnil:2015,Kurata:2016,Rashmi:19}.
 NN-based NLUs have shown remarkable performance but have millions of parameters. They are data hungry, requiring thousands of labeled examples as training data in order to achieve reasonable accuracy \cite{Jerome:2017,Mesnil:2015,Kurata:2016}. This hinders the training and deployment of models for new domains, especially production domains where intents/slots are added over time. 

Zero-shot learning \cite{xian2018zero} enables models to output class labels which have not been seen during training \cite{Bapna:2017}. 
Zero-shot slot labeling can be performed as a sequential labeling task  - see how slot labels are defined in the Traditional vs. Zero-shot setup in Figure \ref{fig:sub2}. In the zero-shot setting, the space of output labels is \textit{Begin}(B), \textit{Inside}(I, to handle multi-word phrases) and \textit{Other/Outside}(O, for words that are not relevant) for each specific slot - as opposed to the space of all possible slot types (such as, \textit{city}, \textit{cuisine}, etc.) in the traditional setting. Prediction is performed by feeding in pairs of (the user's sentence, slot description) for each slot type as input. In the example, the slot description for \textit{year} is passed as input to the model along with the user's sentence. The model then predicts B/I/O labels for individual words/tokens in the user's sentence that may correspond to the \textit{year} slot type. The token, \textit{2014}, is predicted as belonging to the slot, \textit{year} (as it received the label, B) while all other tokens (\textit{Play, music, from}) in the sentence were assigned to O (implying that they do not belong to \textit{year}). This process is repeated for every slot type on the same user's sentence. The individual I/O/B predictions for each slot type in the user's sentence are then aggregated to obtain the final prediction to match the output in the traditional setting.  
Lee et. al., \shortcite{Lee:2018} use slot descriptions to bootstrap to new slots. 
Shah et. al., \shortcite{Shah:2019} utilize both the slot description and a few exemplar slot values as opposed to just using slot descriptions. 

\begin{figure}
\centering
\includegraphics[width=1\linewidth]{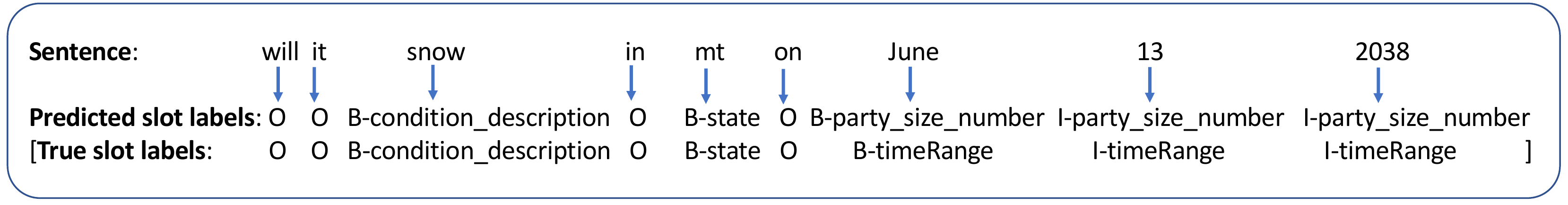}
\caption{Predicted slot labels using the baseline.}
\label{fig:sub1}
\end{figure}

Previous zero-shot approaches have either been applied for intent detection \cite{liu-etal-2019-reconstructing, xia2018zeroshot, Siddique:2021, williams2019zero} or slot labeling \cite{Bapna:2017,Shah:2019} but not both. As a result, these approaches do not consider dependencies between slot labels and intents. Additionally, previously proposed zero-shot slot-labeling approaches \cite{Bapna:2017,Shah:2019} do not capture dependencies between slot labels. To understand the implications of these, consider an example in Figure \ref{fig:sub1} from the popular SNIPS dataset \cite{Alice:2018}.

The presence of \textit{condition\_description} and \textit{state} slot labels are strong indicators that the numbers (\textit{13}, \textit{2038}) provided correspond to \textit{timeRange}. However, prior approaches tend to predict out of context slots, in this case \textit{party\_size\_number} is predicted even though the input sentence does not belong to the \textit{BookRestaurant} intent.
Such inconsistencies can be avoided by capturing (1) dependencies between intents and slot labels and (2) dependencies between slot labels. We incorporate such dependencies using  global constraints in our model and a form of beam search decoding. The predicted \textit{GetWeather} intent can provide additional support for \textit{timeRange}.
In this paper, we propose an architecture (Zero-Shot learning setup in Figure \ref{fig:sub2}) and novel training and inference schemes to jointly predict both intents as well as slot labels, all of which extend to the zero-shot setting for new intents and slots. 
The contributions of this paper are as follows:
\setlist{nolistsep}
\begin{itemize}[noitemsep, leftmargin=*]
    \item We develop zero-shot learning approaches to jointly predict intents and slot labels (Section \ref{sec:method}). The approaches capture dependencies (1) between intents and slot labels and (2) between slot labels, using learned global constraints.
    \item We show that sentence-level representations of input utterances and descriptions of short intents from pre-trained language models (e.g.  BERT \cite{devlin2018bert}), do not lie in a comparable space (Section \ref{sec:problems}). 
    This leads to inaccurate predictions on unseen intents (Section \ref{sec:results}). We show how to learn translations between these spaces within our joint intent-slot model to fix this.
    \item We show that fine-tuning BERT improves traditional intent and slot classification, but hurts zero shot performance. To overcome this limitation, we describe a sequential training procedure that first trains a translation model for sentence and word embeddings and then fine tunes the language model. 
    \item Along the way we also mention important tricks (e.g. how to use negative sampling) that seem to be required to get improvements. 
\end{itemize}
We now describe the experimental set up and datasets used to evaluate our proposed approach and baselines.

\section{Experimental Setup}
We demonstrate results on the SNIPS dataset \cite{Alice:2018}, commonly used for evaluating NLU and is especially appropriate for evaluating zero-shot learning as the data covers multiple domains. The SNIPS data has 7 intent types and 72 slot labels. The training set contains 13,084 utterances, 700 utterances in the test set and development set.  
Similar to test settings used in \cite{Shah:2019, Bapna:2017}, for every target intent, we train a model on utterances from the remaining intents and evaluate on the test set that includes the `missing' or `target' intent. In contrast to the previously proposed approaches that only predict presence of slot labels, we also predict the presence of intents. 
We used the BertAdam optimizer \cite{adam} with warmup ratio of 0.1, with maximum number of epochs set to 20. For ablation models M0-M5 below, the optimal learning rates were varied between 5e-03 to 5e-06
 , and final value chosen based on the performance of the models on a validation set. The reported scores are averaged over 3 random seeds.
Preliminary experiments had unsatisfactory performance. To enable learning, how we sample negative and positive examples for intents and slots is critical. To enable learning from different combinations of intents and slots, 
we sample one example from each of the following (1) an intent is present and a slot label is present (2) an intent is present but a slot label is absent (3) an intent is absent but a slot label is present (4) both intent and slot label are absent. We report accuracy for intent and f1 for slot labeling performance using CONLL's evaluation script \cite{conll}.
\begin{figure}
     \centering
     \begin{subfigure}[b]{0.46\textwidth}
         \centering
         \includegraphics[width=\textwidth]{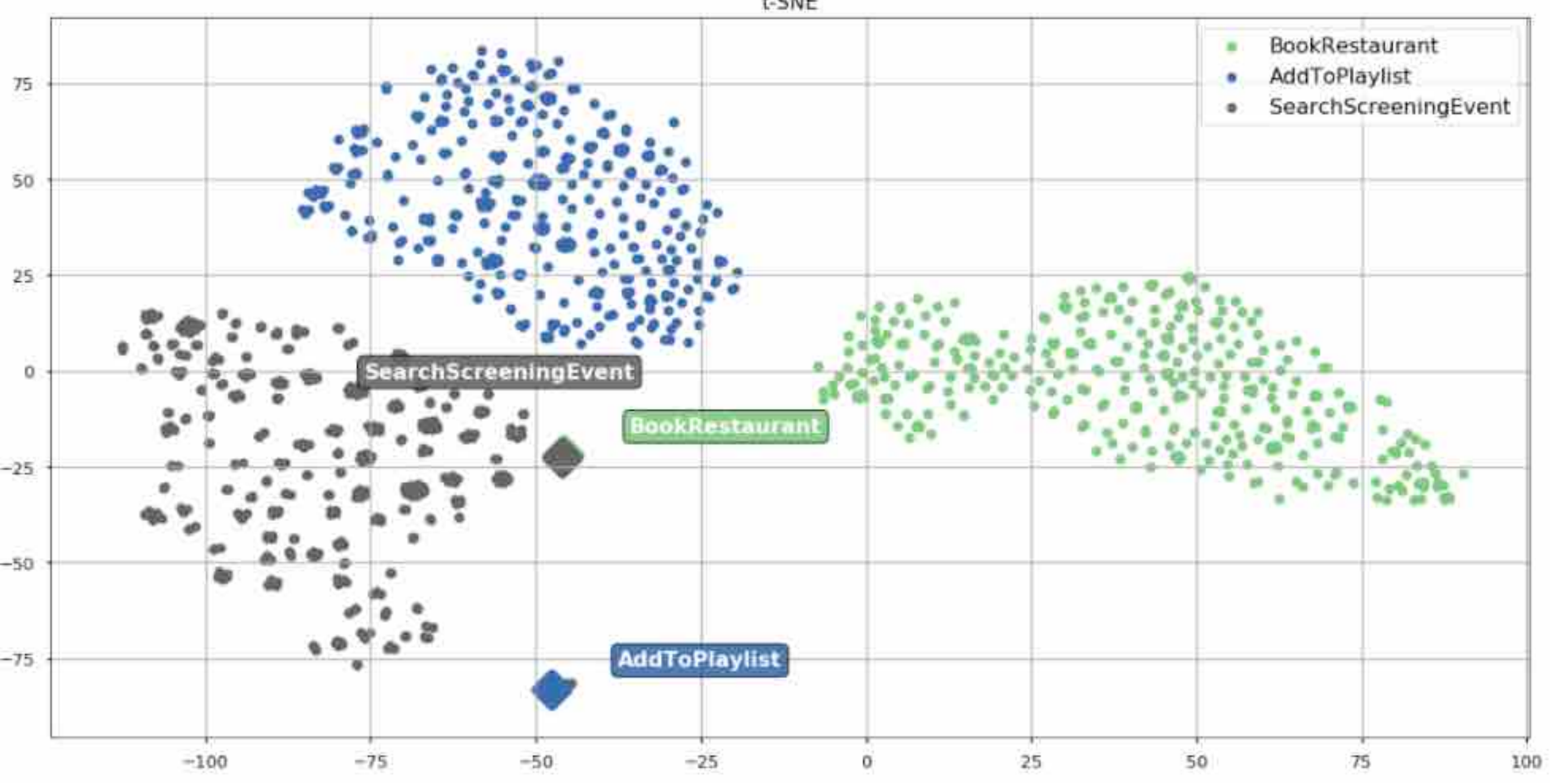} 
         \caption{
         }
         \label{fig1}
     \end{subfigure} 
     \begin{subfigure}[b]{0.46\textwidth}
         \centering
         \includegraphics[width=\textwidth]{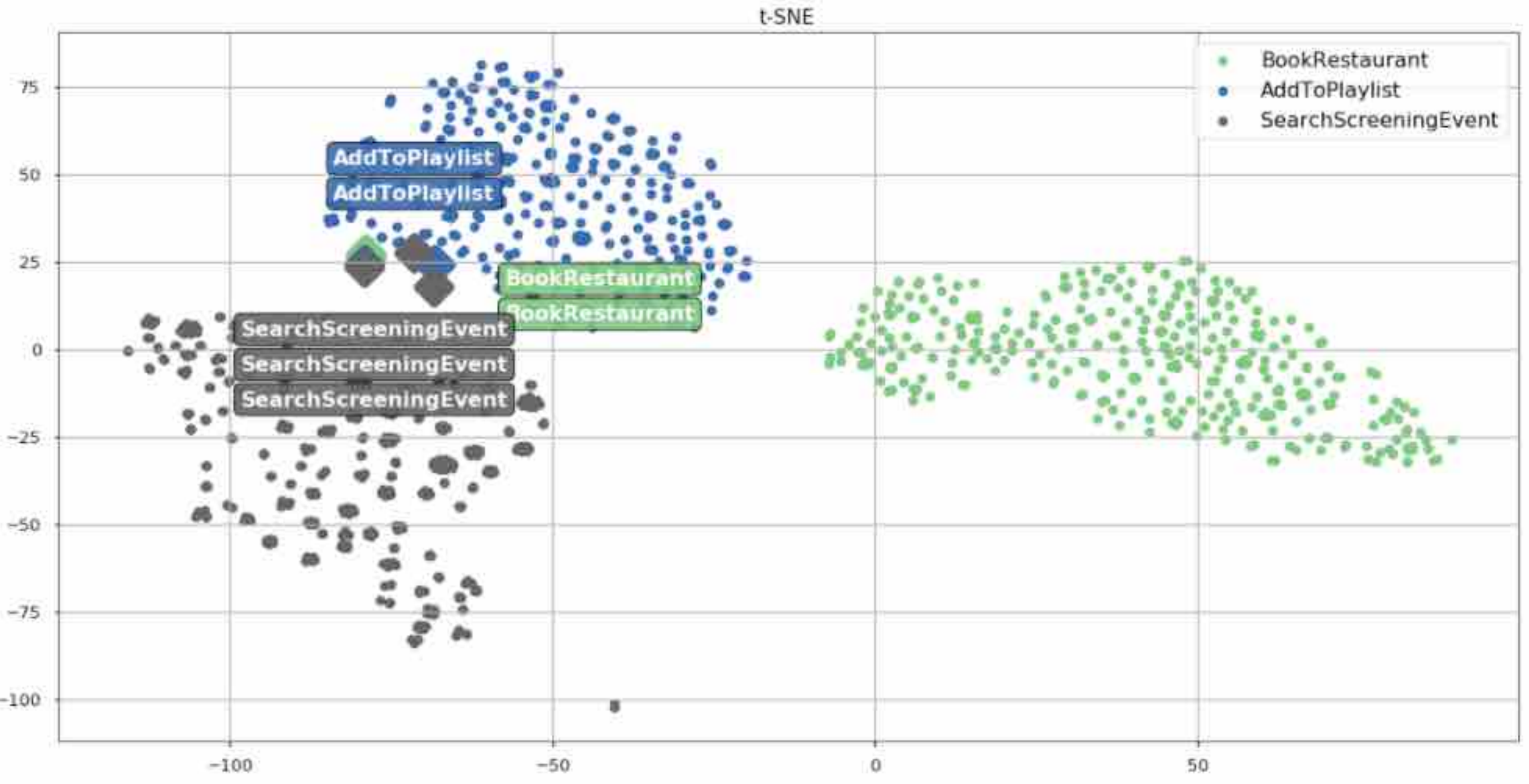}
         \caption{}
         \label{fig2}
     \end{subfigure}
     \hfill
     \begin{subfigure}[b]{0.46\textwidth}
         \centering
         \includegraphics[width=\textwidth]{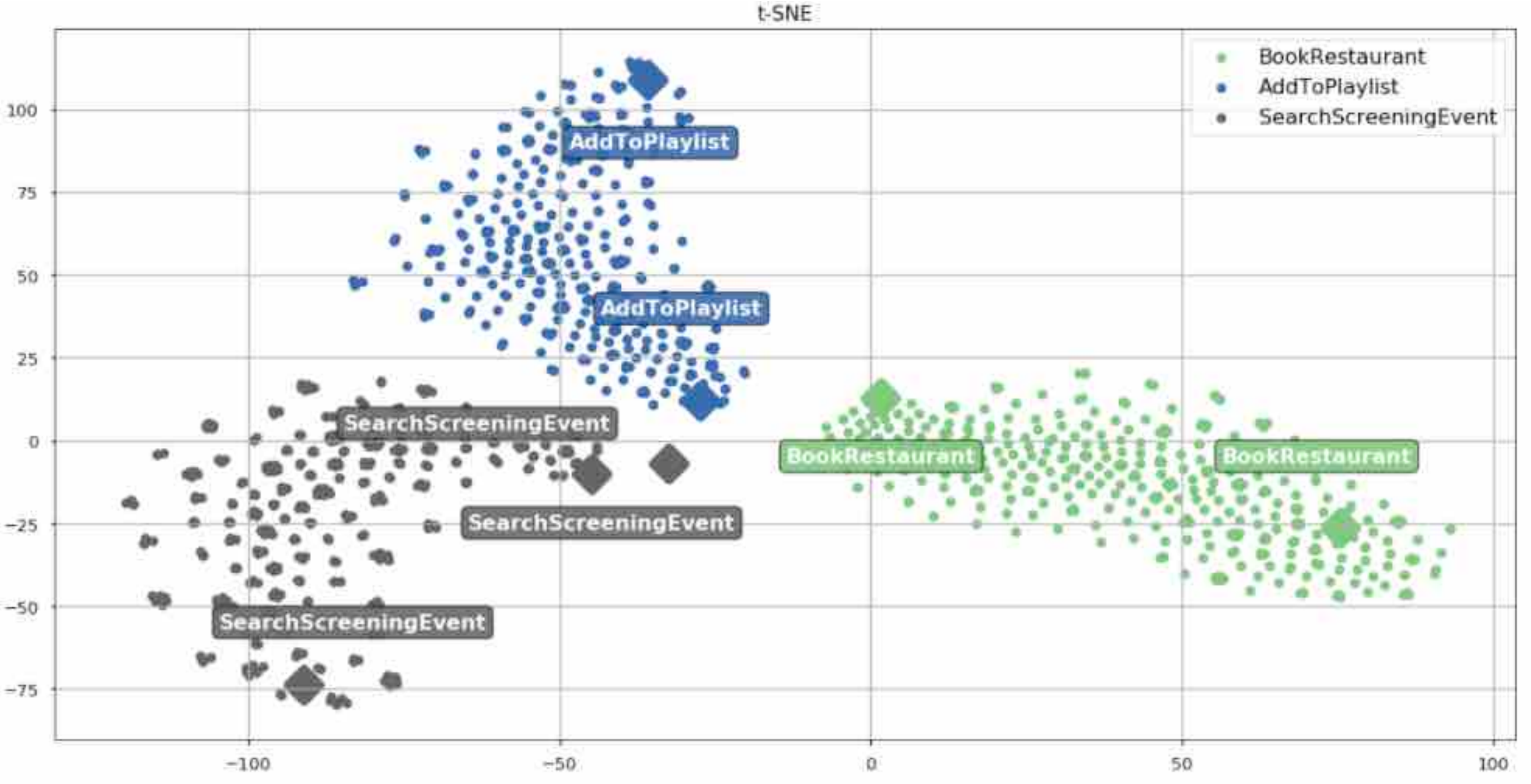}
         \caption{}
         \label{fig3}
     \end{subfigure}
        \caption{(a) S-level-reps of intent descriptions and input sentences (b) S-level-reps of input sentences and W-level-reps of intent descriptions (c) proposed representations for input sentence ($g^a$) and intent descriptions}
        \label{fig:threeplots}
\end{figure}

\section{Issues with sentence representations}
\label{sec:problems}
Following prior work in slot labeling in zero-shot settings \cite{Shah:2019,Bapna:2017,Lee:2018} we leverage slot labels to create descriptions, e.g., \textit{GetWeather} is converted to \textit{get weather} and \textit{condition\_temperature} is converted to \textit{condition temperature}. This is reasonable since intent and slot names typically convey information on what they represent.  
We leverage BERT \cite{devlin2018bert} to obtain sentence-level ($\mathbf{CLS}$/S-level-reps) and token-level (W-level-reps) representations, due to its superior performance on many NLP tasks, especially on traditional joint intent and slot labeling, where sufficient number of examples are available for all intents and slot types \cite{chen2019bert}. Since zero-shot settings rely on an encoding that is a function of the input sentence and the descriptions, we further analyzed the representation spaces of both the sentence and token-level representations obtained via BERT. 

We found that it is not useful to compare S-level-reps of input sentences and S-level-reps of very short intent descriptions (Figure \ref{fig:threeplots}(a), we consider only 3 intents for this plot for comprehensible visualization). Another option would be to compare the word(W)-level-reps of these short intent descriptions with S-level-reps of the input sentences. However, as Figure \ref{fig:threeplots}(b) illustrates, W-level-reps and S-level-reps do not lie in a comparable space. To overcome this problem, we propose using attention mechanisms on the W-level-reps of the intent descriptions, together with translation layers that can map between spaces (Figure \ref{fig:threeplots}(c)). 




\section{Proposed Method}
\label{sec:method}
Consider a user's sentence with $\mathbf{T}$ tokens. The corresponding BERT S-level-reps and W-level-reps (of size $dim$) are passed through translation layers, $Z_S$ and $Z_W$ respectively, to obtain, $\{x_{\mathbf{CLS}}, x_i \in \textbf{R}^{dim}, i\in[1,\mathbf{T}]\}$. $x_\mathbf{CLS}$ stands for the sentence representation and $x_i$ is the token-level representation for the token $i$. $Z_S$ and $Z_W$ are composed of two fully connected feed forward layers. Similarly, we obtain translated representations $\{a_{\mathbf{CLS}}, a_q \in R^{dim}, q \in [1,\textbf{Q}]\}$ for $\textbf{Q}$ intent description tokens, $\{b_{\mathbf{CLS}}, b_s \in \textbf{R}^{dim}, s\in[1, \textbf{S}]$\} for $\textbf{S}$ slot description tokens, $\{e^k_{\mathbf{CLS}},e_k^n \in \textbf{R}^{dim}, n\in[1, \textbf{N}_k]\}$ for each of $\textbf{K}$ examples per slot, with $\textbf{N}_k$ tokens.

\subsection{Slot prediction}
We perform mean pooling of the translated representations and compute an attention weighted representation of all the $\mathbf{K}$ slot examples for each token in the user's utterance $u$, resulting in a $\mathbf{R}^{dim}$ representation. 
\begin{eqnarray}
\label{eqn:1}
 & & e_k=\frac{1}{\mathbf{N}_k} \sum_{n=1}^{\mathbf{N}_k} e^n_k,\ \ k\in[1,\mathbf{K}]\nonumber \\
& & \alpha^i_k = \mathbf{softmax}(x_i W_s e_k) \nonumber \\
& & e^s_i = \sum_{k=1}^{\mathbf{K}} \alpha^i_k \odot e_k  
\end{eqnarray}
$\odot$ denotes dot product and $\oplus$ denotes concatenation.
The resulting representation at each token position $i$, $e^s_i$ from (\ref{eqn:1}), is then concatenated with $x_i$, $b_{\mathbf{CLS}}$ and $a_{\mathbf{CLS}}$ and sent through a $\mathbf{biLSTM}$ \cite{Sepp:97} layer, followed by a $\mathbf{softmax}$ layer to obtain predictions for the slot labels.
\begin{eqnarray}
\label{eqn:2}
&&d_i=\mathbf{biLSTM}(\{x_i \oplus b_{\mathbf{CLS}} \oplus a_{\mathbf{CLS}} \oplus e^s_i\}) \nonumber \\
&& y_i=\mathbf{softmax}(W_y d_i+\textrm{bias}_y)
\end{eqnarray}
\subsection{Intent prediction} 
We then perform max pooling to obtain $g^a \in \mathbf{R}^{dim}$, followed by a dense and $\mathbf{sigmoid}$ layer to obtain, $z^a$, corresponding to the likelihood that intent $a$ is present (we overload $a$).
\begin{eqnarray}
\label{eqn:3}
& & \beta^i_q = \mathbf{softmax}(\{x_i W_p a_q\}) \nonumber \\
& & g_i^a = \sum_{q=1}^{\mathbf{Q}} \beta^i_{q} \odot a_q \nonumber \\
& & g^a = \mathbf{max}_i(g^a_i)\nonumber \\
& & z^a=\mathbf{sigmoid}\ (\ \mathbf{Dense}(a_\mathbf{CLS} \oplus g^a)\ )
\end{eqnarray}
We also compare this with a variation that uses sentence representations of the input sentence and the intent description where,
\begin{equation}
\label{eqn:4}
g^a = x_{\mathbf{CLS}} \odot a_{\mathbf{CLS}}
\end{equation}
We are essentially interested in the distance between learned embeddings of intents, slots and utterances. The euclidean distance between two vectors $x$ and $y$ (\cite{snell2017prototypical}), can often be well approximated by networks with inner products, but often not by networks where the inputs are concatenated. In particular, if $|x|_2$ approximately proportional $(\approxprop$) $|y|_2$, $|x-y|^2 \approxprop - (x\odot y)$. In our experiments, we found that element-wise multiplication layers outperformed concatenation layers.

\subsection{Global Slot Constraint:}
We learn a global constraint vector that captures dependencies between slot labels- dependencies previously ignored by vanilla zero-shot slot label prediction. We consider descriptions of all slot labels of the intent. Say, $\{b_1,..b_\mathbf{L}\}$ represent translated vectors of all slot label descriptions that belong to the intent (e.g., \textit{artist}, \textit{album}, \textit{genre}, etc. for the intent \textit{PlayMusic}), with each intent containing $\mathbf{L}$ slot labels and each slot description containing $\mathbf{M}_l$ tokens.
\begin{eqnarray}
\label{eqn:5}
& & b_l=\frac{1}{\mathbf{M}_l} \sum_{m=1}^{\mathbf{M}_l} b_m^l, l\in[1,\mathbf{L}] \nonumber\\
& &\gamma^i_l = \mathbf{softmax}(\{x_i W_r b_l\}) \nonumber \\
& &\gamma_l = \textbf{max}_i(\gamma^i_l)
\end{eqnarray}
$\gamma \in \mathbf{R}^\mathbf{L}$ is concatenated in (\ref{eqn:2}) and (\ref{eqn:3}), resulting in:
\begin{eqnarray}
&&\hspace{-0.3in}d_i=\mathbf{biLSTM}(\{x_i \oplus b_\mathbf{CLS} \oplus a_\mathbf{CLS} \oplus \gamma \oplus e_i^s\}) \nonumber \\
&&\hspace{-0.3in}z^a=\mathbf{sigmoid}(\ \mathbf{Dense}(a_\mathbf{CLS} \oplus \gamma \oplus g^a)\ )
\end{eqnarray}
We use categorical cross entropy loss for slots and binary cross entropy loss for predicting the presence of intents. The losses are combined using a weight that is learned along with the rest of the parameters of the model \cite{kendall:2018}.

\begin{table*}[ht!]
  \begin{center}
    \begin{tabular}{c||c|c|c|c|c||c|c|c|c|c|c|c|c|c}
    \hline
      Metric$\rightarrow$ & \multicolumn{5}{c||}{Intent Accuracy} & \multicolumn{9}{c}{Slot F1 score} \\
    \hline 
      Model$\rightarrow$ & M1 & M2 & M3 & M4 & M5 & CT & ZT & XT & M0 & M1 & M2 & M3 & M4 & M5\\
      Target Intent$\downarrow$ & & & & & & & & & & & & & & \\
     \hline 
    GetWeather &  83.0 & 83.4 & 84.4 & 88.0 & 88.6 & 63.5& 60.7& 66.0 & 85.8 & 86.0 & 85.9 & 81.3 & 87.7 & 90.9\\ 
    \hline 
    BookRestaurant &  85.0 & 86.0 & 87.6 & 91.9 & 93.8 & 45.7& 46.6& 48.6 & 79.4 & 82.3 & 87.2 & 79.2 & 83.3 & 84.8 \\ 
    \hline 
    PlayMusic & 86.5 & 86.3 & 86.3 & 87.9 & 89.3 & 28.7 & 30.1 & 33.8 & 83.4 & 80.4 & 82.0 & 85.1 & 86.0 & 88.3 \\ 
    \hline 
    AddToPlaylist &  79.6 & 80.6 & 79.0 & 80.6 & 81.0 & 53.3 & 46.8 & 55.2 & 83.7 & 85.8 & 78.8 & 83.0 & 83.4 & 84.7 \\ 
    \hline 
    SearchCreativeWork & 79.5 & 85.0 & 89.0 & 87.1 & 87.1 & 24.7 & 26.7 & 26.2 & 85.3 & 85.8 & 84.6 & 83.6 & 83.5 & 85.8 \\ 
    \hline 
    SearchScreeningEvent &  82.7 & 84.5 & 85.3 & 86.7 & 87.7 & 23.7 & 19.7 & 25.5 & 82.7 &  86.6 & 84.0 & 81.5 & 85.0 & 86.6 \\ 
    \hline 
    RateBook &  88.2 & 88.3 & 88.0 & 88.3 & 90.7 & 24.5 & 31.0 & 28.5 & 76.4 & 78.2 & 81.8 & 75.1 & 79.0 & 83.3 \\ 
    \end{tabular}
\caption{Comparing baselines (CT \cite{Bapna:2017}, ZT \cite{Lee:2018}, XT \cite{Shah:2019}) with models M0-5, in terms of intent accuracy and slot F1 scores.} 
\label{tab:1}
  \end{center}
\end{table*}

\section{Results}
\label{sec:results}
Table \ref{tab:1} shows results for the following:
\setlist{nolistsep}
\begin{itemize}[noitemsep, leftmargin=*]
\item \textbf{M0}: slot labeling only, using eqn (\ref{eqn:2}),
\item \textbf{M1}: joint intent and slot labeling using W-level-reps in eqn (\ref{eqn:4}),
\item \textbf{M2}: joint intent and slot labeling with S-level-reps as in eqn (\ref{eqn:3}),
\item \textbf{M3}: Joint intent and slot labeling with eqn (\ref{eqn:3}) but without fine tuning the BERT layer, 
\item \textbf{M4}: A model initialized using M3, with a second training stage with the translation parameters ($Z_S$ and $Z_W$) fixed, 
\item \textbf{M5}: beam search decoding on M4's predictions. M0, M1 and M2 do not use the translation layers $Z_S$ and $Z_W$. 
\item We also report scores from prior work on zero shot learning for slot labeling \textbf{CT} \cite{Bapna:2017}, \textbf{ZT} \cite{Lee:2018}, \textbf{XT} \cite{Shah:2019}.
\end{itemize}

Comparing M0 with M1-M5, it is clear that slot labeling can be improved by incorporating information about intents. Hence, a joint model of intents and slot labels is better than predicting slot labels independent of intents.
Further analysis of the intent predictions of M1 revealed that most of the unseen target intents had zero (at most 2) matches with the ground truth intent. This implies that the S-level-reps of BERT are not suitable for predicting unseen intents as also seen with the t-SNE plot in Figure \ref{fig1}.
\begin{table}[ht!]
\begin{center}
\begin{tabular}{ c|c|c|c|c|c|c|c } 
 \hline
 Intent$\rightarrow$ & GW & BR & PM & ATP & SCW & SSE & RB \\ 
  \hline
 \%TPR & 51.0 & 81.5 & 20.9 & 33.0 & 30.0 & 36.4 & 45.0 \\
 \%FDR & 0.0 & 13.8 & 0.0 & 0.0 & 13.9 & 12.3 & 2.7\\
 \hline
\end{tabular}
\caption{\%TPR and \%FDR using M5. 
GW: GetWeather, BR: BookRestaurant, PM: PlayMusic, ATP: AddToPlaylist, SCW: SearchCreativeWork, SSE: SearchScreeningEvent, RB: RateBook.}
\label{tab:2}
\end{center}
\end{table}

Comparing M2 and M3, we notice a drop in performance for some intents and slots even though M3 was able to find a substantial number of matches of the target intent, perhaps due to the absence of BERT fine-tuning \cite{Peters:2019} in M3. To take advantage of BERT fine-tuning and yet obtain the performance of M2 on target intents, we perform two stage training where the model (M4) is initialized using M3 and trained with $Z_S$ and $Z_W$ parameters fixed. The resulting model, M4, performs significantly better than M2 on intent prediction and in most cases on slot labeling.

\subsection{Beam Search Decoding with Constraints}
Dependencies between intents and slot labels are already captured in the outputs of our model, since we jointly model intents and slot labels and have a global slot constraint $\gamma$ that captures dependencies between slot labels. 
To further enforce dependencies between slots and intents at inference time, we perform a variant of beam search \cite{Russell} as a post processing step that finds tags for the input sequence by only considering legal paths. 
 At every word position, the algorithm keeps track of top $\textbf{V}=3$ (beam width) legal paths that respect intent and slot constraints. The decoded path in Figure \ref{fig:sub1} would be considered illegal as \textit{party\_size\_number} and \textit{condition\_description} cannot co-occur. Each utterance in the SNIPS dataset belongs to a single intent only. 
 Beam search on multiple intents \cite{Rashmi:19} will be explored as future work. 

 
 The top $3$ highest scoring intents are considered as the initial paths and extended via beam search. A beam search matrix of size $\textbf{T}$ rows and $2*\textbf{S}+1$ columns (where $2*\textbf{S}+1$ is the total number of slot labels including $\text{B}-$ and $\text{I}-$ for every slot label and $\text{O}$) is created by averaging the slot probabilities $p(\text{slot}_j|w_i)$ assigned by the model to each token $w_i$ across all intents. 
 \begin{equation}
 p\text{(O}|w_i)= \max (\epsilon,1-\sum_{j=1}^{2S} p(\text{slot}_j|w_i))
 \end{equation}
 We use $\epsilon = 10^{-7}$.
Beam search on M4's predictions (shown as M5) shows further improvements in slot labeling and intent detection.

We also report True Positive Rate (\%TPR, predicted target intent is the same as the ground-truth target intent) and False Discovery Rate (\%FDR, a target intent is predicted when ground-truth has a different intent) on target intents for M5 in Table \ref{tab:2}. 
We see that M5 found a significant number of target intent matches, with room for improvement for certain intents. 

\section{Conclusion and Future work}
\label{sec:conclusion}
We explored strategies to predict unseen intents and slots in a zero-shot learning setup where we jointly modeled both slot labeling and intent prediction. We showed the importance of capturing dependencies between intents and slot labels using learned global constraints. Through experimentation, we showed how fine-tuning BERT hurt zero shot performance. To overcome this limitation, we proposed a sequential training procedure to first train a translation model for sentence and word embeddings and then fine tune the language model.
As seen in Table \ref{tab:2}, although the proposed models found target intent matches, the performance is still low. We will continue to explore strategies to further improve the performance of NLUs on unseen intents.

\bibliographystyle{ACM-Reference-Format}
\bibliography{paper}


\end{document}